\documentclass[reprint, amsmath, amssymb, aps, superscriptaddress]{revtex4-1}

\usepackage{physics} 
\usepackage{mathtools}
\usepackage{hyperref}
\hypersetup{
    colorlinks=true,
    linkcolor=magenta,
    filecolor=black,      
    urlcolor=magenta,
    citecolor=magenta
}

\newcommand{\Qeta}{\textbf{Q}_{\eta}}
\newcommand{\Qxi}{\textbf{Q}_{\xi}}
\newcommand{\Qxit}{\tilde{\textbf{Q}}_{\xi}}
\newcommand{\Cxy}{\textbf{C}_{XY}}
\newcommand{\Cxty}{\textbf{C}_{\tilde{X}Y}}
\newcommand{\II}{\textbf{I}}
\newcommand{\XX}{\boldsymbol{X}}
\newcommand{\XXt}{\tilde{\boldsymbol{X}}}
\newcommand{\EX}{\langle \boldsymbol{X} \rangle}
\newcommand{\DX}{\Delta\boldsymbol{X}}
\newcommand{\DXt}{\Delta\tilde{\boldsymbol{X}}}

\newcommand{\DD}{\boldsymbol{D}}
\newcommand{\VV}{\boldsymbol{V}}

\begin{document}
\title{A Note on Noisy Reservoir Computation}
\author{Anthony M. Polloreno}
\thanks{ampolloreno@gmail.com}
\author{Reuben R. W. Wang}
\affiliation{Department  of  Physics,  University  of  Colorado,  Boulder  CO  80309,  USA}
\affiliation{JILA, University of Colorado, Boulder, Colorado 80309, USA}
\author{Nikolas A. Tezak}
\thanks{nikolas.tezak@gmail.com, work done outside of OpenAI}
\affiliation{OpenAI, San Francisco, California, USA}
\date{\today}

\begin{abstract}
In this note we extend the definition of the Information Processing Capacity (IPC) by Dambre et al \cite{Dambre2012} to include the effects of stochastic reservoir dynamics. We quantify the degradation of the IPC in the presence of this noise.\\
{[1]} Dambre et al. \emph{Scientific Reports} 2, 514, (2012). \\
\end{abstract}

\maketitle
\section{Reservoir Computation}\label{sec:reservoirs}

In the traditional framework of a classical reservoir computer,  \cite{Tanaka2018_NN} one considers a dynamical system observed at discrete time-steps $t = 0, 1, 2, \ldots$, with internal states $\boldsymbol{s}(t) \in \mathbb{R}^n$, driven by inputs $\boldsymbol{U}(t) \in \mathbb{R}^d$. The reservoir dynamics are then encompassed by the dynamical update rule
\begin{align} \label{eq:classical_update}
    \boldsymbol{s}(t) = \boldsymbol{F}\big( \boldsymbol{s}(t-1), \boldsymbol{U}(t) \big), 
\end{align}
from which $m$ outputs $x_k(t) \in \mathbb{R}$ with $k \leq m \leq n$, can be constructed from the reservoir as
\begin{align}
    x_k(t) = \boldsymbol{G}\big( \boldsymbol{s}(t), \boldsymbol{U}(t) \big).
\end{align}
These outputs can then be utilized to approximate a target function $y(t) \in \mathbb{R}$, via a supervised learning scheme.

Specifically, we first define the concatenated $h$-step sequence of recent inputs $\boldsymbol{U}^{-h}(t) = [ \boldsymbol{U}(t - h + 1), \boldsymbol{U}(t - h + 2), \ldots, \boldsymbol{U}(t)]$. While we may use the reservoir to learn a function of time, the reservoir outputs themselves can be approximated by maps $x_k^h:\boldsymbol{U}^{-h}(t)\mapsto\mathbb{R}$. In particular, this is because we require that the reservoir satisfies the fading memory property \cite{Dambre2012}. A dynamical system has fading memory if, for all $\epsilon > 0$, there exists a positive integer $h_0\in\mathbb{N}$, such that for all $h>h_0$, for all initial conditions,
and for all sufficiently long initialization times $T'>h$, the $x_k(t)$ at any time $t \geq 0$ are well-approximated by functions $x_k^h$:
\begin{equation}
\mathbb{E}(x_k(t) - x_k^h[\boldsymbol{U}^{-h}(t)])^2 < \epsilon
\end{equation}
where the expectation is taken over the $t + T'$ previous inputs.

The signals $x_k(t)$ can used to construct estimators (denoted with hats) of $y(t)$ 
\begin{align}
    \hat{y}_{\boldsymbol{w}}(t) 
    = 
    \boldsymbol{w}^T \boldsymbol{X}(t),
\end{align}
where $\boldsymbol{w}$ is a vector of weights, and $\boldsymbol{X}(t)$ is the vector whose entries are $x_k(t)$.

Characterizing the learning capabilities of a dynamical system requires analyzing it in a learning-task independent manner. We do so by utilizing a quantity known as the information processing capacity (IPC) \cite{Dambre2012}. We first define the \textit{capacity} to reconstruct some arbitrary signal $y$ as
\begin{align}
    C_T[y] = 1 - \min_{\boldsymbol{w}} \frac{ \langle ( \hat{y}_{\boldsymbol{w}} - y)^2 \rangle_T }{ \langle y^2 \rangle_T },
\end{align}
where $\langle \chi \rangle_T = \frac{1}{T} \sum_{t=1}^T \chi(t)$ denotes a time average. Then for a complete and countably infinite set of basis functions $\{ y_1, y_2, \ldots \}$ for the Hilbert space of functions with fading memory \cite{Dambre2012}, the IPC of a dynamical system is given as
\begin{align}\label{eq:ipc}
    {\rm IPC} = \lim_{D \to \infty} \lim_{T \to \infty} \sum_{\ell}^{D} C_T[y_{\ell}] \leq n.
\end{align}
Intuitively, the IPC can be thought of as a normalized measure of the size of the subspace of functions learnable by the dynamical system. Here the basis set should be ordered in some sense by the complexity (e.g., the polynomial degree and the memory time $h$) of the basis functions.
The numerical estimate of the IPC requires defining a probability measure over all possible sequences of inputs (thus promoting $\boldsymbol{U}(t)$ to a random process with measure $\xi$) and truncating the basis set at some sufficiently high, but finite cutoff.

\section{Noisy Reservoir Computation}\label{sec:noisy_reservoirs}
In this section we extend the ideas from Sec.~\ref{sec:reservoirs} and the formula for the IPC in \cite{Dambre2012} to describe the performance of a reservoir in the presence of noise. The internal state and readouts $(\boldsymbol{s}(t), \boldsymbol{X}(t))$, thus evolve stochastically, even for a fixed input sequence. Formally, what this means is that in the following, $\boldsymbol{X}(t)$ and $\boldsymbol{U}(t)$ are random variables with probability measures $\eta$ and $\xi$. Furthermore, because the state of the reservoir is a function of the input to the reservoir, we have that in general $\boldsymbol{X}(t)$ is conditioned on all past $\boldsymbol{U}(t)$.

In what follows, we use the overline $\overline{\boldsymbol{A}}$ to denote the expectation value of some variable $\boldsymbol{A}$ over the input signal distribution $\eta$. We use $\langle \boldsymbol{A} \rangle$ to denote expectation values of $\boldsymbol{A}$ over the reservoir noise distribution $\xi$. We have otherwise suppressed the notational dependence on ${\boldsymbol U}(t)$ in this analysis.

We find that the IPC is 
\begin{align} \label{eq:noisy_ipc}
    {\rm IPC}  
    & \le  
    \Tr( (\II + \Qxit)^{-1}) \\
    & = 
    \sum_{k=1}^n \frac{1}{1 + \tilde{\sigma}_k^2} \le n,
\end{align}
where $\II$ is the identity matrix and $\Qxit$ measures the normalized reservoir output noise covariance. This is by design a positive (semi-)definite matrix whose eigenvalues we denote as $\{\tilde{\sigma}_k^2:\, k=1,\dots,n\}$ and which are the generalized reservoir noise variances.
Thus ${\rm IPC} = n$ only occurs when $\Qxit = 0$ is the trivial matrix, demonstrating that noise strictly decreases the achievable IPC of a reservoir computer of given (output) size $n$.
In the following derivation of the inequality ~\eqref{eq:noisy_ipc} we assume that the average reservoir outputs $\langle \boldsymbol{X} \rangle$ (i.e, averaged over the reservoir noise) satisfy 
\begin{align} \label{eq:full_signal_rank}
 \Qeta \coloneqq \overline{\EX \EX^T} > 0.   
\end{align}
We will designate this assumption as the full signal rank condition. We note, however, that this only simplifies the analysis, and that the result is correct even when the condition does not hold. We revisit this case at the very end of our discussion.

\subsection{Proof}
We limit our proof to the finite case of a truncated basis set (of size $D$). 
A common quantity to consider in any regression analysis is the expected squared error, $J$, between the predictions made by the model and the data. Thus we start by defining the expected squared reconstruction error

\begin{align}\label{eq:squared_error}
    J 
    &=\overline{ \left \langle \sum_{\ell=1}^D\sum_{k=1}^n(y_\ell(t) - x_k(t)w_{k\ell})^2\right\rangle}\nonumber\\
    &=\overline{ \left \langle \sum_{\ell=1}^D\sum_{k=1}^n(y_\ell(t) - x_k(t)w_{k\ell})(y_\ell(t) - w_{k\ell}x_k(t))\right\rangle}\nonumber\\
    &= 
    \overline{ \langle (\boldsymbol{Y}^T - \boldsymbol{X}^T\textbf{W})(\boldsymbol{Y} - \textbf{W}^T\boldsymbol{X}) \rangle } \nonumber\\
    &= \Tr \overline{ \langle (\boldsymbol{Y} - \textbf{W}^T\boldsymbol{X}) (\boldsymbol{Y}^T - \boldsymbol{X}^T\textbf{W})\rangle },
\end{align}
where here vectors are column vectors, and transposes are row vectors. In particular, $\boldsymbol{Y} \in \mathbb{R}^{D \times 1}$, ${\bf W} \in \mathbb{R}^{m \times D}$ are the vector of estimated signals and weight matrix, respectively, and $\boldsymbol{X} \in \mathbb{R}^{m \times 1}$ is defined as before. We have chosen the convention for $\textbf{W}^T$ so that ${\bf W}^T \boldsymbol{X}$ is an estimator of the target functions.  

In this notation we have suppressed and will continue to suppress the dependence on time, so that $\boldsymbol{Y} \coloneqq \boldsymbol{Y}({\boldsymbol U}^{-h}(t))$ are target functions of the finite-sequence inputs ${\boldsymbol U}^{-h}(t)$, and will be approximated in this scheme by linear functions of the measured outputs $\boldsymbol{X} \coloneqq \boldsymbol{X}(t)$. That is, the goal is to find a weight matrix ${\bf W}$ such that ${\bf W}^T \boldsymbol{X}\approx \boldsymbol{Y}$. For computing the IPC, we will take $\boldsymbol{Y}({\boldsymbol U}^{-h}(t))$ to constitute a complete basis of functions on the finite input sequences, as in Eq.~\eqref{eq:ipc}.  
Differentiating Eq.~\eqref{eq:squared_error} with respect to ${\bf W}$ then gives
\begin{align}
    \nabla_{\bf{W}} J 
    &= \nabla_{\bf{W}} \overline{ \langle \boldsymbol{X}^T {\bf W} {\bf W}^T \boldsymbol{X} - \boldsymbol{Y}^T {\bf W}^T \boldsymbol{X} - \boldsymbol{X}^T {\bf W} \boldsymbol{Y} \rangle } \nonumber\\
    &= \nabla_{\bf{W}} \Tr( \overline{ \langle \boldsymbol{X}^T {\bf W} {\bf W}^T \boldsymbol{X} - 2 \boldsymbol{X} \boldsymbol{Y}^T {\bf W}^T \rangle } ) \nonumber\\
    &= 2  \overline{\langle\boldsymbol{X} \boldsymbol{X}^T}\rangle {\bf W} -  2\overline{\langle \boldsymbol{X} \boldsymbol{Y}^T\rangle},
\end{align}
allowing us to solve for $\bf{W}$ viia the first order optimality condition $\nabla_{\bf{W}} J=0$ as
\begin{align}
    \textbf{W}_*  = \langle \overline{\boldsymbol{X} \boldsymbol{X}^T} \rangle^+  \langle \overline{\boldsymbol{X} \boldsymbol{Y}^T}\rangle
\end{align}
where $A^+$ denotes the Moore-Penrose pseudo-inverse of $A$. Under the full signal rank condition \eqref{eq:full_signal_rank} this is a proper inverse.

Defining the reservoir output noise as
\begin{equation}
\DX \coloneqq \XX - \EX,
\end{equation}
allows decomposing the output into its deterministic and its noise part $\XX = \EX + \DX$.
With this, we find
\begin{align}
    \overline{\langle \XX \XX^T\rangle }
    & = 
    \underbrace{\overline{\EX \EX^T}}_{\Qeta} 
    + 
    \underbrace{\langle \overline{\DX \DX^T}\rangle}_{\Qxi} 
    \\ \nonumber
    &\quad + \overline{\EX \underbrace{\langle \DX^T \rangle}_0} + \overline{\underbrace{\langle \DX \rangle}_0\EX ^T} \\
     & = \Qeta + \Qxi,
\end{align}
i.e., we can decompose this second moment matrix into the second moment matrix of the deterministic output signal $\Qeta$ and the noise covariance $\Qxi$.

Under the full signal rank condition \eqref{eq:full_signal_rank}, we can perform a spectral decomposition $\overline{\EX \EX^T} = \VV \DD \VV^T$ with positive definite, diagonal matrix $\DD$ and an orthogonal matrix $\VV$ and factor the second moment matrix as
\begin{align}
    \overline{\langle \XX \XX^T\rangle} = \VV \DD^{\frac12} \left(\II + \Qxit\right) \DD^{\frac12}\VV^T
\end{align}
where we have defined
\begin{align}
    \Qxit \coloneqq \DD^{-\frac12} \VV^T \Qxi \VV \DD^{-\frac12} = \overline{\langle \DXt\DXt^T\rangle}.
\end{align}
$\XXt$ in turn is related to the above diagonalization of the deterministic second moment matrix $\overline{\EX \EX^T}$, corresponding to a basis change for the outputs
\begin{align}
    \XXt = \DD^{-\frac12} \VV^T \XX \Leftrightarrow \XX = \VV \DD^{\frac12} \XXt.
\end{align}
We can use this same transform to re-express the overlap matrix
\begin{align}
    \Cxy & = \overline{\langle \boldsymbol{X} \boldsymbol{Y}^T\rangle} = \overline{\langle \boldsymbol{X}\rangle \boldsymbol{Y}^T} \\ \nonumber
    & = \VV \DD^{\frac12} \overline{\langle \tilde{\boldsymbol{X}}\rangle \boldsymbol{Y}^T} =\VV \DD^{\frac12} \Cxty.
\end{align}
Substituting $\textbf{W}_*$ back into $J$ and transforming to the normalized outputs gives 
\begin{align}
    J({\bf W}_*)
    &=
    \Tr \overline{ \langle \boldsymbol{Y} \boldsymbol{Y}^T - 2 \boldsymbol{Y} \boldsymbol{X}^T {\bf W}_* + {\bf W}_*^T \boldsymbol{X} \boldsymbol{X}^T {\bf W}_* \rangle } \nonumber\\
    &=
    \Tr( 
    \overline{ \langle \boldsymbol{Y} \boldsymbol{Y}^T \rangle }
    -
     \Cxty^T \left(\II + \Qxit\right)^{-1} \Cxty
    ) \nonumber \\
    & = D - \Tr( \left(\II + \Qxit\right)^{-1} \underbrace{\Cxty \Cxty^T}_{\le \II}) \ge D - n
\end{align}
We note that the trace in the last line is over the reservoir output indices $k=1, 2, \dots, n$ whereas the previous lines have traces over the basis set $l=1, 2, \dots, D$.
Finally, our claim follows from the Cauchy-Schwarz inequality and the fact that $\Tr A B \ge 0$ for positive semi-definite $A, B \ge 0$. 
Also note that under the full signal rank condition we have $\lim_{D, T\to\infty} \Cxty \Cxty^T \to \II$, where in typical physicist fashion we ignore the subtleties around how to carefully take the infinite limits.

For a simple example, consider a deterministic reservoir with orthonormal outputs, i.e. $\Qeta = \II$. By introducing Gaussian noise on the outputs with covariance matrix $\bf{\Sigma}$, the IPC is bounded exactly as in Eq.~\eqref{eq:noisy_ipc}, giving
\begin{equation}
{\rm IPC} \leq \sum^{n}_{k=1}\frac{1}{1 + \sigma_k^2},
\end{equation}
where $\sigma_k^2$ are the eigenvalues of $\bf{\Sigma}$, since $\Qeta = \II$ implies $\Qxit = \Qxi = \bf{\Sigma}.$
While we have specified that the noise is on the outputs of the reservoir, our derivation shows that the result is the same independent of whether the noise is added to the outputs or is internal to the reservoir.

Finally, note that this proof can easily be extended to the case without the full signal rank condition \eqref{eq:full_signal_rank}. E.g., if the second signal moment has only rank $\tilde{n} < n$, then it requires replacing $\DD^{-\frac12}$ with its pseudo-inverse version $(\DD^{\frac12})^+$ and some special care needs to be taken when canceling $(\DD^{\frac12})^+\DD^{\frac12} = \bf{I}_{\tilde{n}}$ where $\bf{I}_{\tilde{n}}$ is a diagonal matrix with $\tilde{n}$ ones on its diagonal (and otherwise only zeros). Also, note that in this case the overlap matrix $\Cxy$ has rank $\le \min{(\tilde{n}, D)}$, which implies that for our optimal weights $\bf{W}_\ast$ we are only sensitive to those output noise contributions that are in the linear span of the expected output signals. In particular, the noise covariance then generalizes to
\begin{align}
    \Qxit \coloneqq (\DD^{\frac12})^+ \VV^T \Qxi \VV (\DD^{\frac12})^+ = \overline{\langle \DXt\DXt^T\rangle}
\end{align}
which may have lower rank than $\Qxi$.
\section{Acknowledgments}
AMP acknowledges funding from a NASA Space Technology Graduate Research Opportunity award. NAT acknowledges helpful discussion with Hideo Mabuchi, Marcus P. da Silva and Hakan Tureci.
\bibliography{main}
\end{document}